%% file: main.tex
\documentclass{article}

\PassOptionsToPackage{numbers,compress}{natbib}

\usepackage[final]{arxiv}

\usepackage[utf8]{inputenc} 
\usepackage[T1]{fontenc}    
\usepackage{hyperref}       
\usepackage{url}            
\usepackage{booktabs}       
\usepackage{amsfonts}       
\usepackage{nicefrac}       
\usepackage{microtype}      
\usepackage{graphicx}

\usepackage{enumitem}
\usepackage{listings}
\usepackage[font=small,labelfont=bf,hypcap=false]{caption} 
\usepackage[newfloat,frozencache,cachedir=.]{minted}
\usepackage{appendix}       

\usepackage{subfig}

\usepackage{inconsolata}
\setlist{leftmargin=10.5mm}
\title{CompressAI:\@ a PyTorch library and evaluation platform for end-to-end
	compression research}

\author{%
  Jean B\'{e}gaint\\
  InterDigital AI Lab\\
  \texttt{jean.begaint@interdigital.com} \\
  \And%
  Fabien Racap\'{e} \\
  InterDigital AI Lab\\
  \texttt{fabien.racape@interdigital.com} \\
  \And%
  Simon Feltman\\
  InterDigital AI Lab\\
  \texttt{simon.feltman@interdigital.com} \\
  \And%
  Akshay Pushparaja\\
  InterDigital AI Lab\\
  \texttt{akshay.pushparaja@interdigital.com} \\
}

\begin{document}

\maketitle

\input{01_abstract}
\input{02_introduction}
\input{03_design_goals}
\input{04_features}
\input{05_results}
\input{06_future_work}
\input{07_acknowledegments}

\bibliographystyle{unsrt}
\bibliography{references}

\input{08_appendix}

\end{document}

%% file: 01_abstract.tex
\begin{abstract}

This paper presents CompressAI, a platform that provides custom operations,
layers, models and tools to research, develop and evaluate end-to-end image and
video compression codecs. In particular, CompressAI includes pre-trained models
and evaluation tools to compare learned methods with traditional codecs.
Multiple models from the state-of-the-art on learned end-to-end compression have
thus been reimplemented in PyTorch and trained from scratch. We also report
objective comparison results using PSNR and MS-SSIM metrics vs. bit-rate, using
the Kodak image dataset as test set. Although this framework currently
implements models for still-picture compression, it is intended to be soon
extended to the video compression domain.

\end{abstract}

%% file: 02_introduction.tex
\section{Introduction}

Following the success of deep neural networks in out-performing most
conventional approaches in computer vision applications, Artificial Neural
Network (ANN) based codecs have recently demonstrated impressive results for
compressing
images~\cite{balle_variational_2018,minnen2020channel,cheng_learned_2020,klopp2018learning,mentzer_high-fidelity_2020}.

Conventional lossy image compression methods like JPEG~\cite{wallace1992jpeg},
JPEG2000~\cite{skodras2001jpeg}, HEVC~\cite{sullivan2012overview} or
AV1~\cite{chen2018overview} or VVC~\cite{vvc} have iteratively improved on a similar coding
scheme: partition images into blocks of pixels, use transform domain to decorrelate
spatial frequencies with linear transforms (\textit{e.g.}: DCT or DWT), perform some predictions
based on neighboring values, quantize the transformed coefficients, and finally
encode the quantized values and the prediction side-information into a bit-stream with
an efficient entropy coder (\textit{e.g.}: CABAC~\cite{marpe2003context}). On the other
hand, ANN-based codecs mostly rely on learned analysis and synthesis non-linear
transforms. Pixel values are mapped to a latent representation via an analysis
transform, the latent is then quantized and (lossless-ly) entropy coded.
Similarly, the decoder consists of an approximate inverse transform, or synthesis
transform, than converts the latent representation back to the pixel domain.

By learning complex non-linear transforms based on convolutional neural networks
(CNN), ANN-based codecs are able to match or outperforms conventional
approaches. The training objective is to minimize the estimated length of the
bitstream while keeping the distortion of the reconstructed image low, compared to
the original content. The distortion can be easily measured with objective or
perceptual metrics like the MSE (mean squared error) or the MS-SSIM (multi-scale
structural similarity)~\cite{wang2003multiscale}. Minimizing the bit-stream size
requires to learn shared probability models between the encoder and decoder
(priors)~\cite{balle_variational_2018,balle_end--end_2017,minnen2020channel},
as well as using relaxation
methods~\cite{agustsson_universally_2020,theis_lossy_2017,balle_end--end_2017,dumas_autoencoder_2018}
to approximate the non-differentiable quantization of the latent values. The
complete encoding-decoding pipeline can be trained \textit{end-to-end} with any
differentiable distortion metrics, which is especially appealing for perceptual
metrics (learned or approximated) or machine-tasks related metrics (for example
image segmentation/classification at very low-bitrates).

Similarly, significant progresses have been reported regarding neural networks
targeting video
compression~\cite{djelouah2019neural,habibian_video_2019,agustsson2020scale}.
Compressing videos is more challenging as reducing temporal redundancies
requires to estimate motion information (such as optical flow) involving larger
networks and multiple-stages training pipelines~\cite{djelouah2019neural}.

Promising results have been achieved with ANN-based codecs for image/video
compression and further improvements can be excepted as better entropy models,
training setups and network architectures are discovered. As such, more research
and experiments are required to improve the performances of learned codecs.
However as this field is relatively new, there is a lack of tooling to
facilitate researchers contributions. The CompressAI platform, presented in this
document, aims to help improve this situation.

\section{Motivation}

The current deep learning ecosystem is mostly dominated by two frameworks:
PyTorch~\cite{paszke_pytorch_nodate} and TensorFlow~\cite{abadi2016tensorflow}.
Discussing the merits, advantages and particulars of one framework over the
other is beyond the scope of this document. However, there is evidence that
PyTorch has seen a major growth in the academic and industrial research circles
over the last years. On the other hand, building end-to-end architectures for
image and video compression from scratch in PyTorch requires a lot of
re-implementation work, as PyTorch does not ship with any custom operations
required for compression (such as entropy bottlenecks or entropy coding tools).
These required components are also mostly absent from the current PyTorch
ecosystem, whereas the TensorFlow framework has an official library for learned
data compression~\footnote{\url{https://github.com/tensorflow/compression/}}.

CompressAI aims to implement the most common operations needed to build deep
neural network architectures for data compression in PyTorch, and to provide
evaluation tools to compare learned methods with traditional codecs. CompressAI
re-implements models from the state-of-the-art on learned image compression.
Pre-trained weights, learned using the Vimeo-90K training
dataset~\cite{xue2019video}, are included for multiple bit-rate points and
quality metrics, which achieve similar performances to reported numbers in the
original papers. A complete research pipeline, from training to performance
evaluation against other learned and conventional codecs, is made possible with
CompressAI\@.

%% file: 03_design_goals.tex
\section{Design}

CompressAI tries to adhere to the original design principles of PyTorch:
\textit{be pythonic}, \textit{put researchers first}, \textit{provide pragmatic
performance} and \textit{worse is better}~\cite{paszke_pytorch_nodate}. As a
PyTorch library, it also follows the code structures and conventions which can
be found in widely used PyTorch libraries (such as
\texttt{TorchVision}\footnote{\url{https://github.com/pytorch/vision}} or
\texttt{Captum}\footnote{\url{https://github.com/pytorch/captum}}).

As a research library, CompressAI aims to follow the naming conventions
introduced in the literature on learned data compression, as to ease the
transition from paper to code. High level APIs to train or run inference on
models do not require much prior knowledge on learned compression or deep
learning. However, specific code implementations relating to the (learned)
compression domain may require to be familiar with the terminologies. This
facilitates adoption and ease of use for researchers.

%% file: 04_features.tex
\section{Features}

\begin{table}[tp]
	\centering%
	\begin{tabular}{ll}
		\toprule
		Model                         & Description                                                                                                   \\
		\midrule
		\textit{bmshj2018-factorized} & Factorized prior \cite{balle_variational_2018}                                                                \\
		\textit{bmshj2018-hyperprior} & Hyperprior \cite{balle_variational_2018}                                                                      \\
		\textit{mbt2018-mean}         & Hyperprior with a Gaussian mixture model \cite{minnen_joint_2018}                                             \\
		\textit{mbt2018}              & Joint autoregressive and hyperprior \cite{minnen_joint_2018}                                                  \\
		\textit{cheng2020-anchor}     & Extension from~\cite{minnen_joint_2018}, residual blocks and sub-pixel deconvolution~\cite{cheng_learned_2020} \\
		\textit{cheng2020-attn}       & Extension from~\textit{cheng2020-anchor} with attention blocks~\cite{cheng_learned_2020} \\
		\bottomrule
	\end{tabular}
	\caption{\label{tab:inc-models}Re-implemented models from the
		state-of-the-art on learned image compression currently available in
		CompressAI\@. Training, fine-tuning, inference and evaluation of the
		models listed in this table are fully supported. The rate-distortion
		performances reported in the original papers have been successfully
		reproduced from scratch (see~\autoref{subsec:comp-publications}).}
\end{table}

\subsection{Building neural networks for end-to-end compression}

One of the most important features of CompressAI is the ability to easily
implement deep neural networks for end-to-end compression. Several
domain-specific layers, operations and modules have been implemented on top of
PyTorch, such as entropy models, quantization operations, color transforms.

Multiple architectures from the state-of-the-art on learned image
compression~\cite{balle_variational_2018,minnen_joint_2018,cheng_learned_2020}
have been re-implemented in PyTorch with the domain specific modules and layers
provided by CompressAI\@. See~\autoref{tab:inc-models} for a complete list and
description. With a few dozen lines of python code, a fully end-to-end network
architecture can be defined as easily as any PyTorch model
(cf.~\autoref{sec:code-examples} for some code examples).

Currently, CompressAI tools and documentation mostly focus on learned image
compression and will soon add support for video compression. However, other
end-to-end compression pipelines could be built using CompressAI, like the compression
of 3D maps or deep features for example.

\subsection{Model zoo}

CompressAI provides pre-trained weights for multiple state-of-the-art
compression networks. These models are available in a model \textit{zoo} and can
be directly downloaded from the CompressAI API\@. Theses pre-trained weights
allow very near reproduction of the results from the original publications, as well as
fine-tuning of models with different metrics or bootstrapping more complex models.

\paragraph{Zoo API}

\begin{listing}[ht]
\begin{minted}[fontsize=\small]{python}

	from compressai.zoo import bmshj2018_factorized

	net = bmshj2018_factorized(quality=1)
	net = bmshj2018_factorized(quality=3, metric='mse')
	net = bmshj2018_factorized(quality=4, metric='mse', pretrained=True)
	net = net.eval()
\end{minted}
\captionof{listing}{Example of the API to import pre-defined models for specific
	quality settings and metrics with or without pre-trained weights.}
\label{lst:example_zoo_api}
\end{listing}

Instantiating a model from pre-trained weights is fast and straightforward (see
example~\ref{lst:example_zoo_api}). Internally, CompressAI leverages the PyTorch
API to download and cache the serialized object models.

As is common in the PyTorch ecosystem, pre-trained models expect input batches
of RGB image tensors of shape (N, 3, H, W), with N being the batch size and (H, W)
the spatial dimensions of the images. The input image data range should be [0,
1], and no prior normalization is to be performed.

Due to the design of the reference models, some constraints need to be
respected: H and W are expected to be at least 64 pixels long. Based on the number of
strided convolutions and deconvolutions for a particular model, users might have
to pad H and W of the input tensors to the adequate dimensions.

Most of the models have different behaviors for their training or evaluation
modes. For example, quantization operations may be performed differently:
uniform noise is usually added to the latent tensor during training, while rounding is
used at the inference stage. Users can switch between modes via
\textit{model.train()} or \textit{model.eval()}.

\paragraph{Training}

\begin{table}[tp]
	\center%
	\begin{tabular}{lllllllll}
		\toprule
		Metric   & 1      & 2      & 3      & 4      & 5      & 6      & 7      & 8      \\
		\midrule
		MSE      & 0.0018 & 0.0035 & 0.0067 & 0.0130 & 0.0250 & 0.0483 & 0.0932 & 0.1800 \\
		MS-SSIM  & 2.40   & 4.58   & 8.73   & 16.64  & 31.73  & 60.50  & 115.37 & 220.00 \\
		\bottomrule
	\end{tabular}
	\caption{\label{tab:lambdas}Lambda values used for training the networks are
	different bit-rates (quality setting from 1 to 8), for the MSE and MS-SSIM metrics}
\end{table}

\begin{table}[tp]
	\center%
	\begin{tabular}{ll}
		\toprule
		Metric & Loss function \\
		\midrule
		MSE & $\mathcal{L} = \lambda * 255^2 * \mathcal{D}_{MSE} + \mathcal{R}$ \\
		MS-SSIM & $\mathcal{L} = \lambda * (1 -\mathcal{D}_{MS-SSIM}) + \mathcal{R}$ \\
		\bottomrule
	\end{tabular}
	\caption{\label{tab:loss_functions}Loss functions used for training the
		networks, with $\mathcal{D}$ the distortion and $\mathcal{R}$ the estimated
	bit-rate.}
\end{table}

Unless specified otherwise, the provided pre-trained networks were trained for
4\--5M steps on $256\times256$ image patches randomly extracted and cropped from
the Vimeo-90K dataset~\cite{xue2019video}.

Models were trained with a batch size of 16 or 32, and an initial learning rate
of 1e-4 for approximately 1\--2M steps. The learning rate is then divided by 2
whenever the evaluation loss reaches a plateau (we use a patience of 20 epochs).
Training usually takes between 4 or 10 days to reach state-of-the-art
performances, depending on the model architecture, the number of channels and
the GPU architecture used.

The loss functions and parameters used for training are respectively reported in
tables~\ref{tab:loss_functions} and~\ref{tab:lambdas}. The number of channels in
the auto-encoder bottleneck varies depending on the targeted bit-rates. The
bottleneck needs to be larger for higher bit-rates. For low bit-rates, below 0.5
bpp, the literature usually recommends using 192 channels for the entropy
bottleneck, and 320 channels for higher bit-rates. CompressAI provides
downloadable weights for most of the pre-defined architectures, pre-trained
using either MSE or MS-SSIM~\cite{wang2003multiscale} (multi-scale structural
similarity), for multiple bit-rates (6 or 8) up to 2bpp.

\subsection{Utilities}

\begin{listing}[]
	\centering%
	\small{}
	\begin{verbatim}
		python -m compressai.utils.find_close av1 cat.png 33 --metric psnr
		python -m compressai.utils.find_close hm cat.png 0.4 --metric bpp
		python -m compressai.utils.find_close vtm cat.png 0.994 --metric ms-ssim
	\end{verbatim}
	\captionof{listing}{Finding the quality parameter to reach the closest
	metric value via a binary search.}
	\label{lst:example_find_close}
\end{listing}

CompressAI ships with some command line utilities that may come in handy when
developing or evaluating learned image compression codecs. The following
tasks can be performed directly from the command line by calling CompressAI
scripts:

\begin{itemize}
	\setlength{\itemsep}{0pt}
	\item evaluating a pre-trained or user-trained model on a dataset of images
	\item evaluating a conventional codec on a dataset of images
	\item finding the right quality parameter to reach a given PSNR or bit-rate
		on a target image (see example in listing ~\ref{lst:example_find_close})
\end{itemize}

\subsection{Benchmarking}

This section exposes the evaluation tools implemented in CompressAI\@. One of
the design goals was to provide simple and efficient tools for comparing
end-to-end methods and traditional codecs. This allows researchers to reproduce
and validate published results, but also to iterate rapidly over research ideas.

Currently, supported quality metrics are the PSNR (peak signal-to-noise ratio)
and the MS-SSIM.

\subsubsection{Learned models}

Rate-distortion performances of learned models can be measured for each of the included
models in CompressAI, see~\autoref{tab:inc-models} for an exhaustive list of
the re-implemented models.

\subsubsection{Traditional codecs}

To facilitate the comparison with traditional codecs, CompressAI includes a simple
python API and command line interface. The most common image and video codecs
are supported, the complete list of supported codecs and their respective
implementations can be found in~\autoref{tab:codecs}.

Default parameters have been chosen to provide fair and comparable results
between conventional codecs and learned ones.

\begin{table}[]
	\begin{tabular}{lll}
		\toprule
		Codec                            & Implementation & URL                                                        \\
		\midrule
		JPEG~\cite{wallace1992jpeg}      & libjpeg        & \url{http://ijg.org/}                                      \\
		JPEG2000~\cite{skodras2001jpeg}  & openjpeg       & \url{http://www.openjpeg.org/}                             \\
		WebP~\cite{webp}                 & libwebp        & \url{https://chromium.googlesource.com/webm/libwebp}       \\
		BPG~\cite{bpg}                   & libbpg         & \url{https://bellard.org/bpg}                              \\
		HEVC~\cite{sullivan2012overview} & HM             & \url{https://hevc.hhi.fraunhofer.de}                       \\
		AV1~\cite{chen2018overview}      & AOM            & \url{https://aomedia.googlesource.com/aom/}                \\
		VVC~\cite{vvc,vtm}                   & VTM            & \url{https://vcgit.hhi.fraunhofer.de/jvet/VVCSoftware_VTM} \\
		\bottomrule
	\end{tabular}
	\caption{\label{tab:codecs}Supported codecs for benchmarking by CompressAI\@.}
\end{table}

At this time, runtime comparisons between traditional methods and ANN-based
methods cannot be reported in an accurate and fair manner. Measuring the
inference efficiency of an ANN-based codec is an active research
topic~\footnote{http://www.compression.cc/}. Neural networks are effectively
designed to run on massively parallel architectures whereas traditional codecs
are typically designed to run on a single CPU core.

\paragraph{Note:} CompressAI does not ship the binaries of the above traditional codecs but
rather provides a common Python interface over the executables, with the
exceptions of JPEG and WebP which are linked by default using the Python Pillow
Image library.

%% file: 05_results.tex
\section{Evaluation}

\subsection{Comparison with originally published results}\label{subsec:comp-publications}

This section exposes the equivalence between the results produced using
CompressAI by retraining state-of-the-art methods from scratch using Vimeo-90K as
training set, and the originally published results.

The following models have been re-implemented in CompressAI\@:
\begin{itemize}
	\item factorized-prior and hyperprior models from Ball\'{e} \textit{et
	al.}~\cite{balle_variational_2018},
	\item hyperprior with non-zero Gaussian means and auto-regressive models
	from Minnen \textit{et al.}~\cite{minnen_joint_2018}.
	\item anchor and self-attention models joint hyperprior models from Cheng
		\textit{et al.}~\cite{cheng_learned_2020}.
\end{itemize}

The pre-trained weights, optimized for the MSE (Mean-Square-Error) metric, can be directly
accessed from the CompressAI API\@. Pre-trained weights optimized for the MS-SSIM
metric are also being added.

\begin{figure}[htb]
	\subfloat[Factorized prior
	model~\cite{balle_variational_2018}~\label{fig:1:m1}]
	{\includegraphics[width=0.50\textwidth]{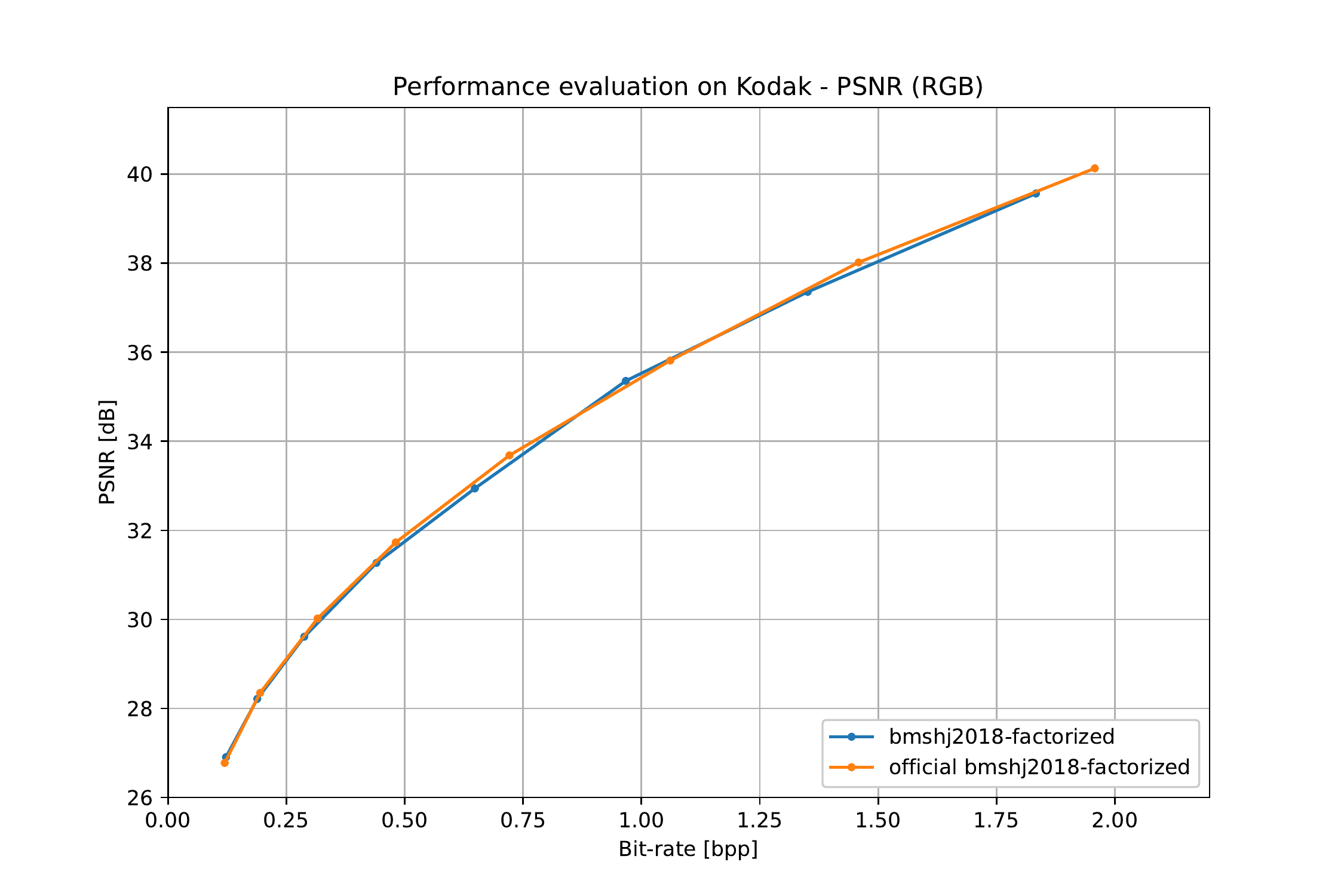}}
	\qquad %
	\subfloat[Hyperprior
	model~\cite{balle_variational_2018}~\label{fig:1:m2}]
	{\includegraphics[width=0.50\textwidth]{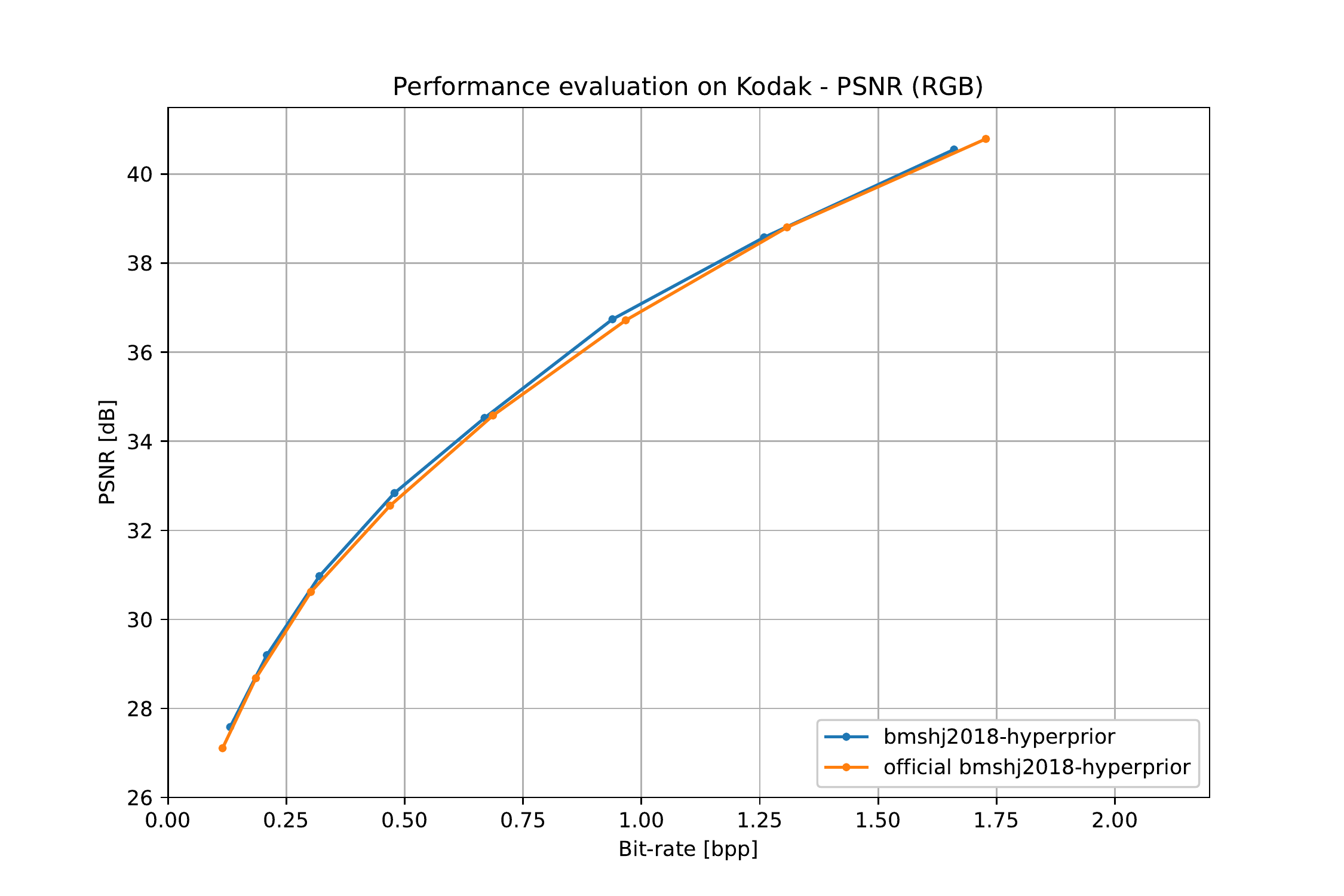}}\\
	\subfloat[Hyperprior model with non-zero mean
	gaussian~\cite{minnen_joint_2018}\label{fig:1:m3}]
	{\includegraphics[width=0.50\textwidth]{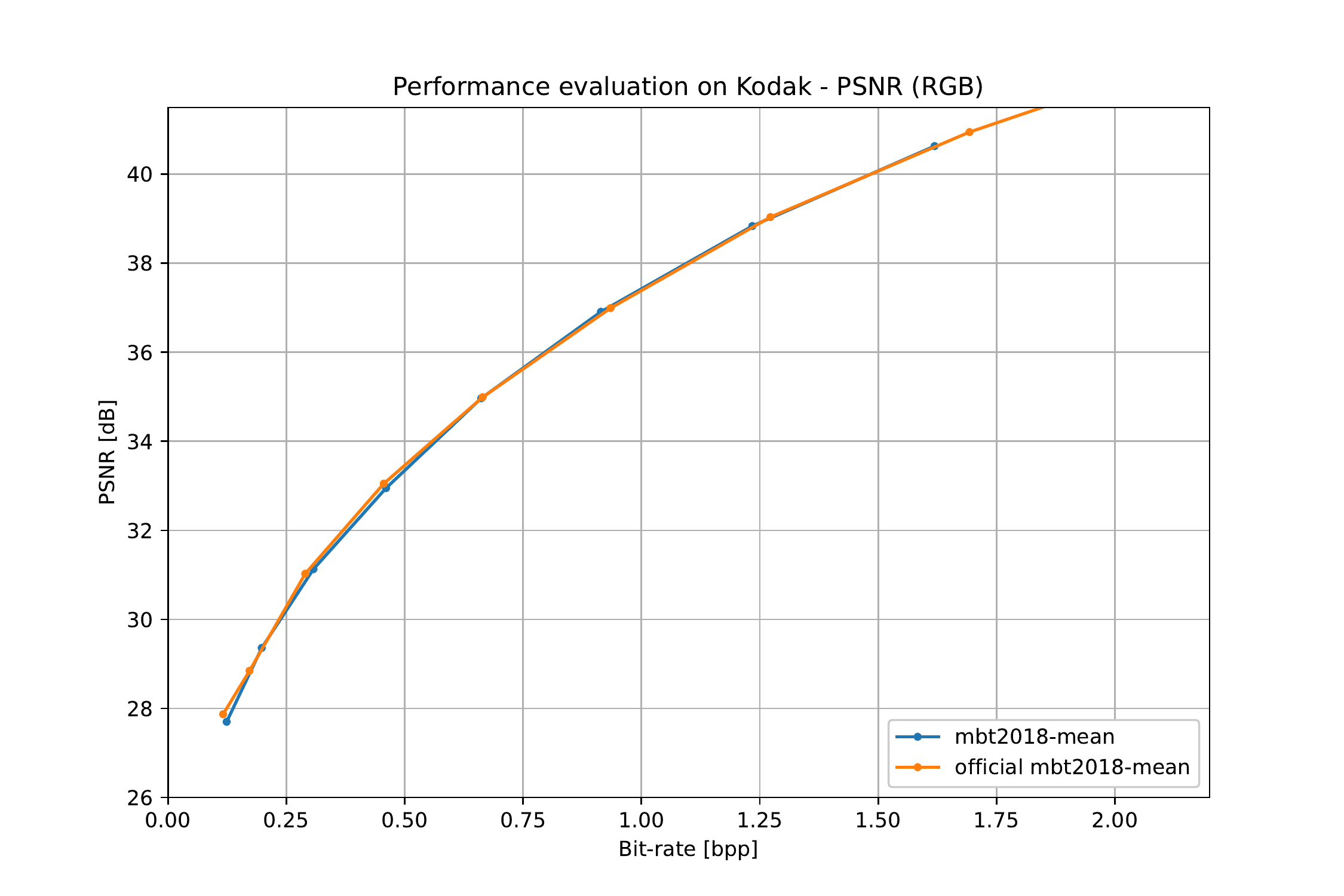}}
	\qquad %
	\subfloat[Joint model~\cite{minnen_joint_2018}\label{fig:1:m4}]
	{\includegraphics[width=0.50\textwidth]{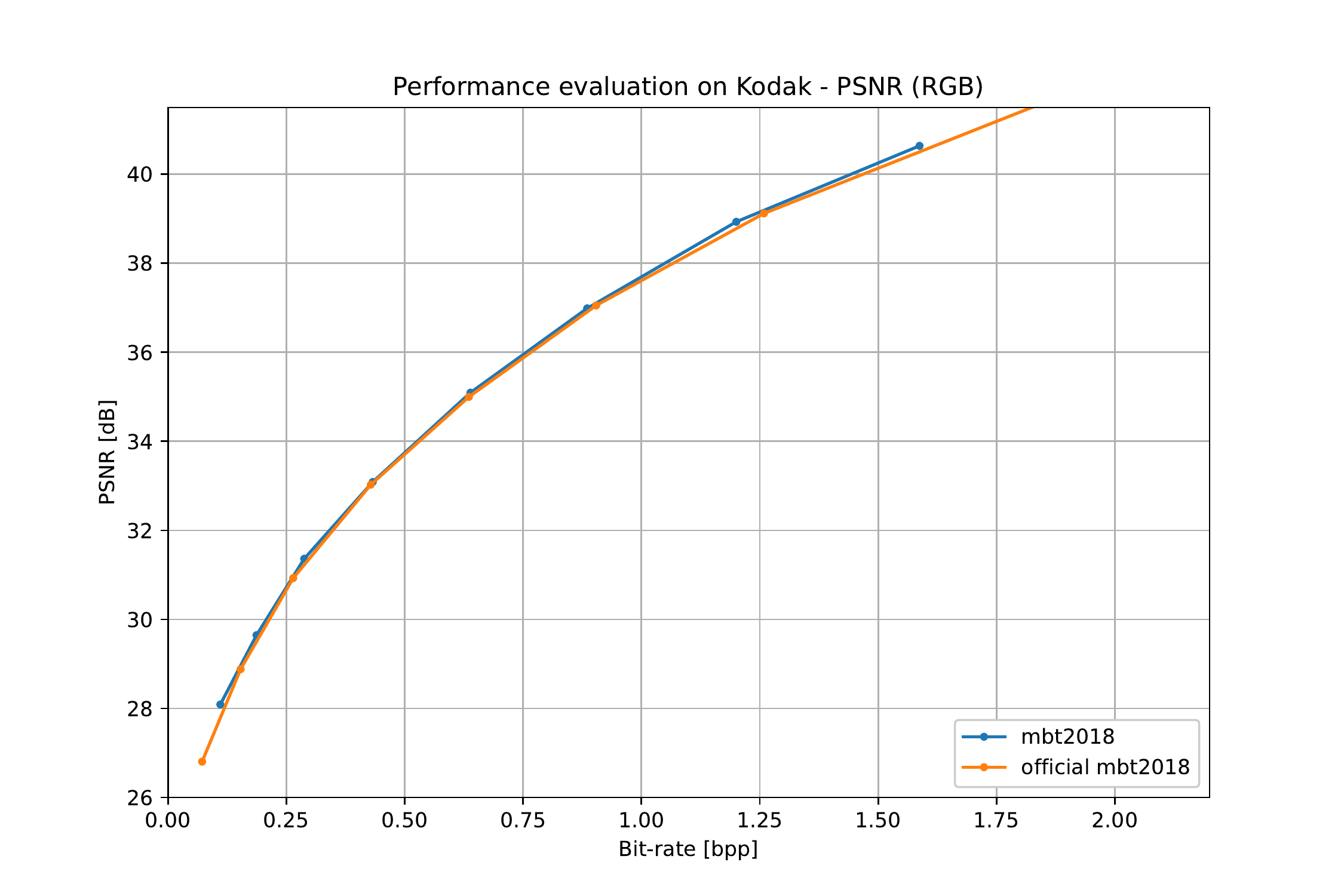}}
	\caption{%
		Performance comparison between models included in CompressAI and results
		from the original publications. The performance is measured in terms of PSNR
		(RGB) vs.\ bit-rate on the Kodak dataset. In each case, the blue curve
		corresponds to the results obtained with CompressAI and the orange curve
		corresponds to the original results reported by the authors. Note that
		these results have been obtained using a different training dataset and
		implementation, highlighting the feasibility of training custom learned
		codecs.\label{fig:1:all}%
	}
\end{figure}

The following graphs in~\autoref{fig:1:all}, which report the average
performance on the Kodak dataset~\cite{eastman_kodak_kodak_nodate}, show that
similar results have been reproduced to those reported in the original
publications. Note that these are actual bit-rates counted on the produced
bit-streams, not the estimated entropy values provided by the networks. The full
encoding/decoding pipeline is implemented within CompressAI\@. However, due to
floating point operations (at the auto-encoder network and probability
estimation levels), reproducibility across different systems or platforms is not
yet achieved. Some publications on the subject have already proposed solutions
(\textit{e.g.}: integer models in~\cite{}), and will be considered in a future
version.

\subsection{Learned codecs versus traditional video codecs}

In this section, we provide a brief objective comparison between learned codecs
and traditional methods. Pre-trained models provided with CompressAI are
compared with HEVC (HM version 16.20), VVC (VTM version 9.1)~\cite{vtm}  and AV1 (version
2.0). HEVC, VVC and AV1 are configured following their default intra mode
configuration and with 8-bit YCbCR 4:4:4 inputs/outputs.

\begin{figure}[]
  \centering
  \includegraphics[width=\textwidth]{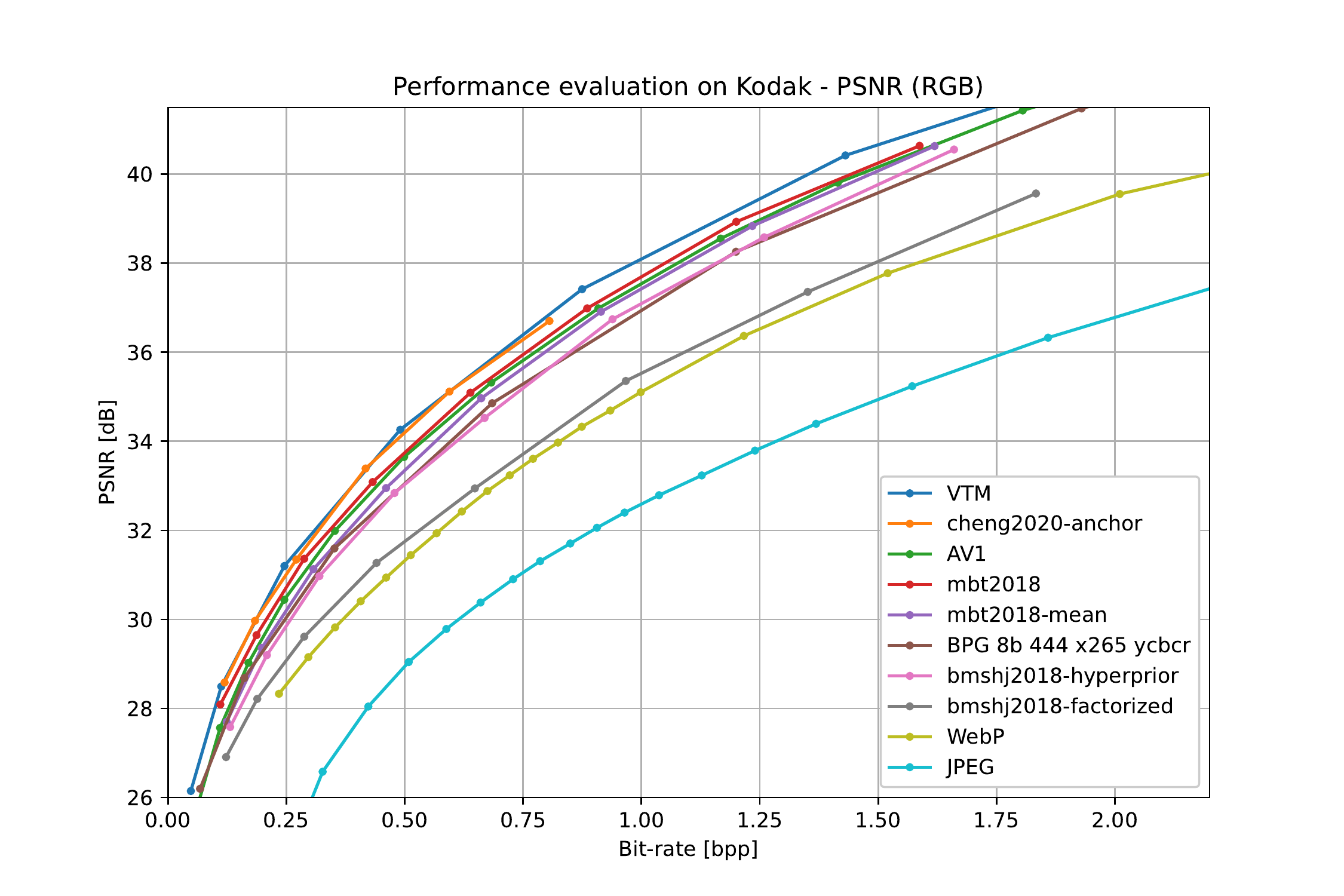}
  \captionof{figure}{\label{fig:results_kodak_small}%
	  Traditional and learned image codecs compared on the Kodak
	  dataset~\cite{eastman_kodak_kodak_nodate}. The PSNR is computed over the
	  RGB channels and aggregated over the whole dataset.
	  See~\autoref{fig:results_kodak} in Appendix~\ref{appendix:rd-curves} for
	  larger rate-distortion curves and a larger set of compression methods.}
\end{figure}

As can be seen in~\autoref{fig:results_kodak_small}, recent works on learned
image compression~\cite{minnen_joint_2018, cheng_learned_2020,
balle_variational_2018} compare favorably with already published standards such
as H.265/HEVC and AV1. The most performing methods are competitive
with the latest ITU/ISO codec H.266/VVC in PSNR at low bit-rates. However, learned
compression frameworks can be directly optimized for complex objective metrics as
long as the metric is differentiable or has a differentiable approximation. This
constitutes a major asset compared to traditional hybrid codecs where it is
difficult to define good strategies for block-based encoder decisions.
CompressAI will soon support different metrics for training and evaluation.
See Appendix~\ref{appendix:rd-curves} for more results with the
MS-SSIM metric.

Besides, learned image and video compression codecs have only started to achieve
competitive results these last 5 years. Considering the success of learned
methods in other image processing and computer visions domains, significant
improvements in compression performance and inference speed can be expected as
more research will be performed on the subject.

\subsection{Adoption}

The first public version of CompressAI was released on GitHub in early June
2020. Since then, we have already noticed adoption from both the industrial and
academic research communities. Multiple MPEG contributions in the Deep Neural
Network for Video Coding (DNNVC) and Video Coding for Machine (VCM) working
groups~\footnote{http://wg11.sc29.org/} were based on CompressAI\@.  Several
academic research groups have also started using CompressAI for their research.

%% file: 06_future_work.tex
\section{Conclusion and future work}

CompressAI currently implements networks for still picture coding and provides
pre-trained weights and tools to compare state-of-the-art models with
traditional image codecs. It reproduces results from the literature and
allows researchers, developers and enthusiasts to train and evaluate their own
neural-network-based codec. 

Several extensions to CompressAI are planned. In the next releases, CompressAI will
include additional models from the literature on learned image compression, and more 
pre-trained weights for perceptual metrics (\textit{e.g.}: MS-SSIM~\cite{wang2003multiscale}). 
One critical extension is to add support for video compression.
Evaluation for low-delay and random-access video coding with traditional codecs,
and end-to-end networks with compressible motion information modules will be
introduced in the next releases. A better compatibility with TorchScript and
ONNX is also being considered.

The platform is made available to the research and open-source communities under the
Apache 2.0 license. We plan to continue supporting and extending CompressAI openly
on GitHub, and we welcome feedback, questions and contributions.

%% file: 07_acknowledegments.tex
\section{Acknowledgements}

The authors would like to thank Chamain Hewa Gamage for thoughtful discussions
and valuable comments on CompressAI\@. The authors would also like to thank the
authors of the TensorFlow Compression
library~\footnote{https://github.com/tensorflow/compression/} for open-sourcing
their code.

%% file: 08_appendix.tex
\newpage
\appendix

\section{Rate-distortion curves}\label{appendix:rd-curves}

\subsection{PSNR on Kodak}

\begin{figure}[htb]
  \centering
  \includegraphics[width=0.9\textwidth]{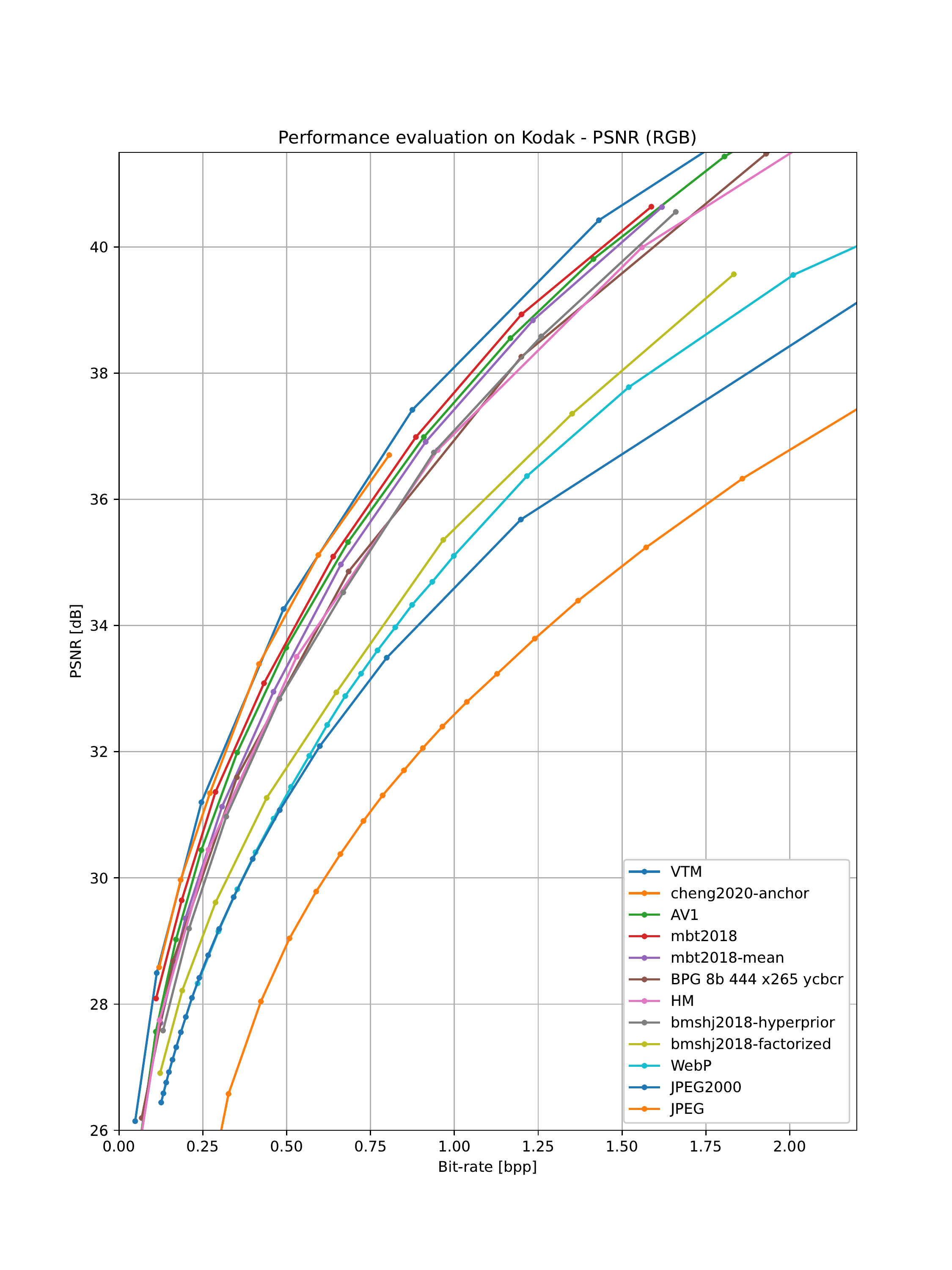}
  \captionof{figure}{\label{fig:results_kodak}%
	  Rate-distortion curves for PSNR measured on the Kodak
	  dataset~\cite{eastman_kodak_kodak_nodate}. JPEG, JPEG 2000 and WebP are
	  largely outperformed by all the learned methods. More recent codecs like
	  HEVC/BPG, AV1 and VVC are also challenged by hyperprior-based
	  methods~\cite{balle_variational_2018,minnen_joint_2018,cheng_learned_2020},
	  which can reach similar or better performances.
  }
\end{figure}

\newpage
\subsection{MS-SSIM on Kodak}
\begin{figure}[htb]
  \includegraphics[width=0.9\textwidth]{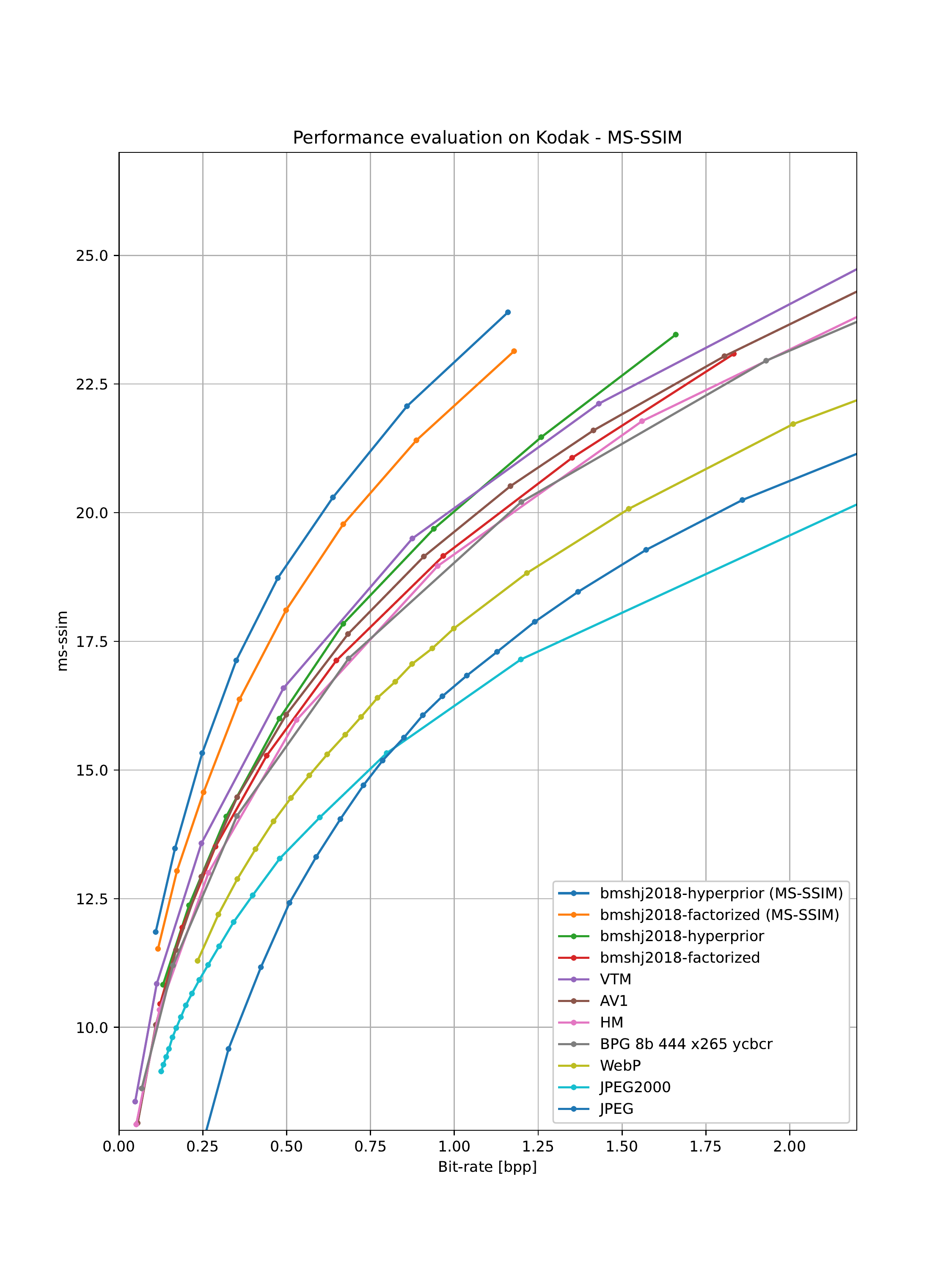}
  \captionof{figure}{\label{fig:results_kodak_ms_ssim}%
	  Rate-distortion curves for MS-SSIM measured on the Kodak
	  dataset~\cite{eastman_kodak_kodak_nodate}. Hyperprior and factorized
	  models from~\cite{balle_variational_2018} are fine-tuned with the MS-SSIM
	  metric. Learned methods significantly outperform traditional methods, even
	  when trained using the MSE performances are competitive or better.
  }
\end{figure}

\newpage
\subsection{PSNR on CLIC Mobile (2020)}
\begin{figure}[htb]
  \includegraphics[width=0.9\textwidth]{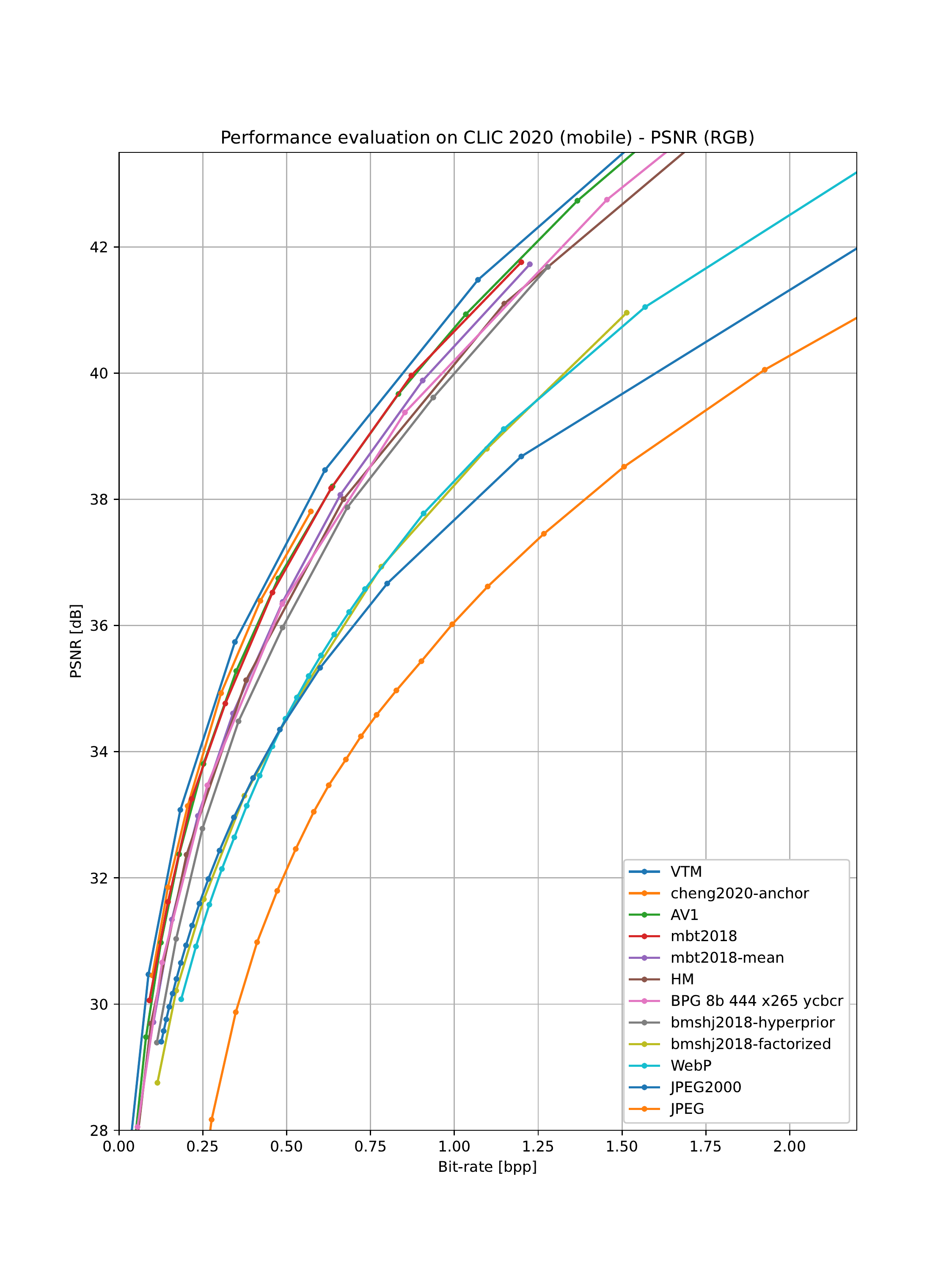}
  \captionof{figure}{\label{fig:results_clic_mobile}%
	  Rate-distortion curves for PSNR measured on the CLIC Mobile
  dataset\cite{CLIC}.}
\end{figure}

\newpage
\subsection{MS-SSIM on CLIC Mobile (2020)}
\begin{figure}[htb]
  \includegraphics[width=0.9\textwidth]{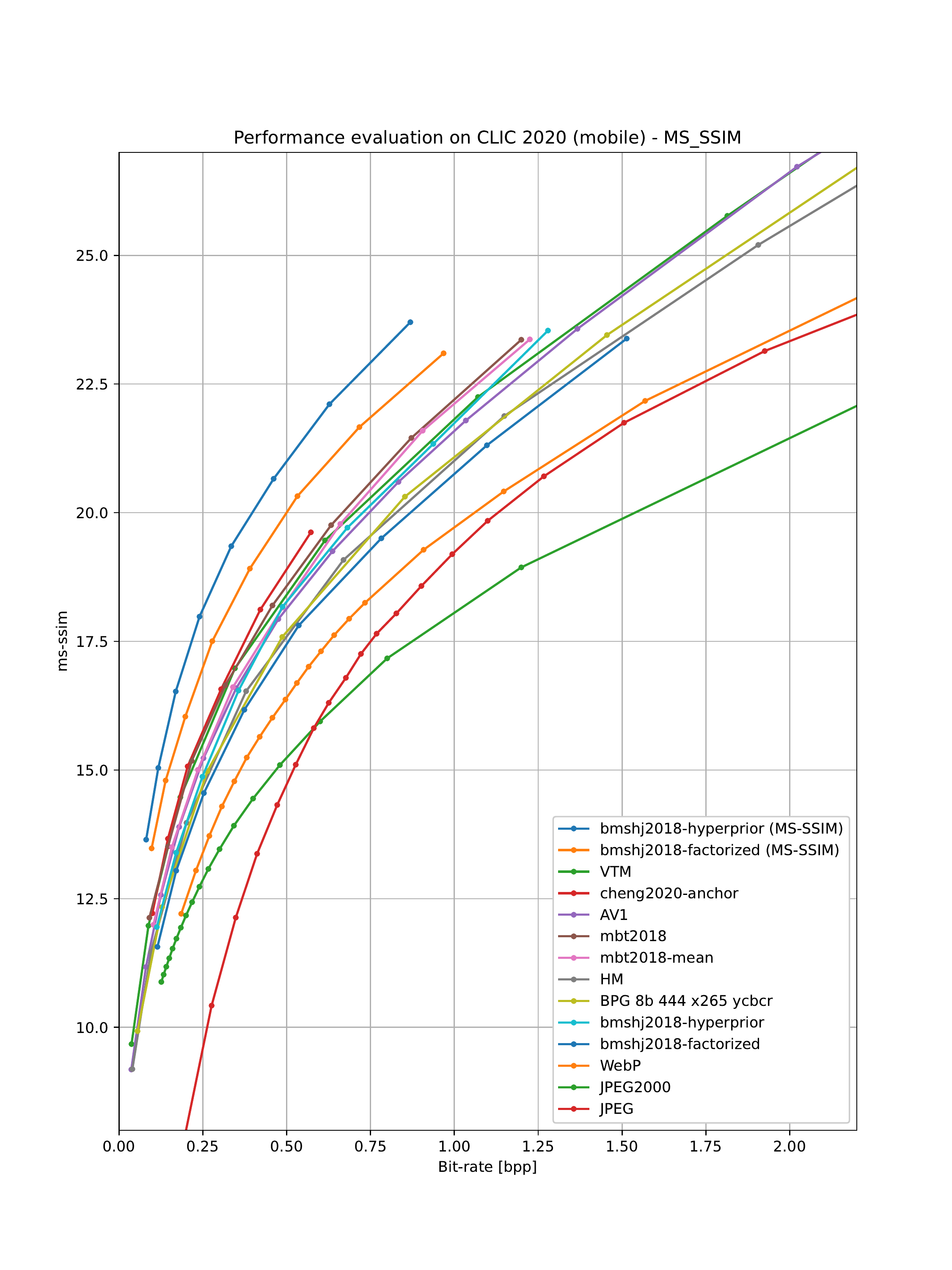}
  \captionof{figure}{\label{fig:results_clic_mobile_ms_ssim}%
	  Rate-distortion curves for MS-SSIM measured on the CLIC Mobile
  dataset\cite{CLIC}.}
\end{figure}

\newpage
\subsection{PSNR on CLIC Pro (2020)}
\begin{figure}[htb]
  \includegraphics[width=0.9\textwidth]{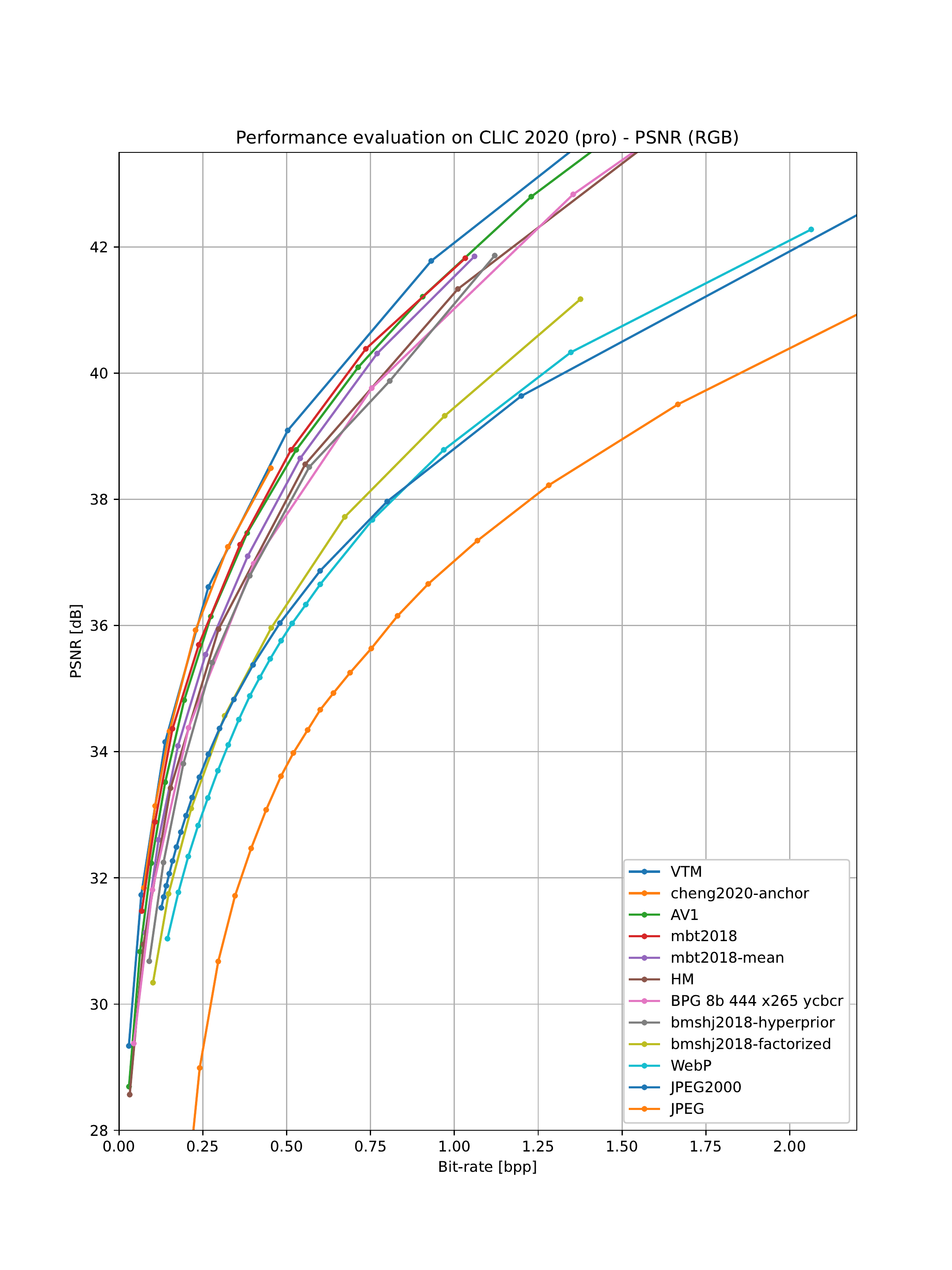}
  \captionof{figure}{\label{fig:results_clic_pro}%
	  Rate-distortion curves for PSNR measured on the CLIC Pro
  dataset\cite{CLIC}.}
\end{figure}

\newpage
\subsection{MS-SSIM on CLIC Pro (2020)}
\begin{figure}[htb]
  \includegraphics[width=0.9\textwidth]{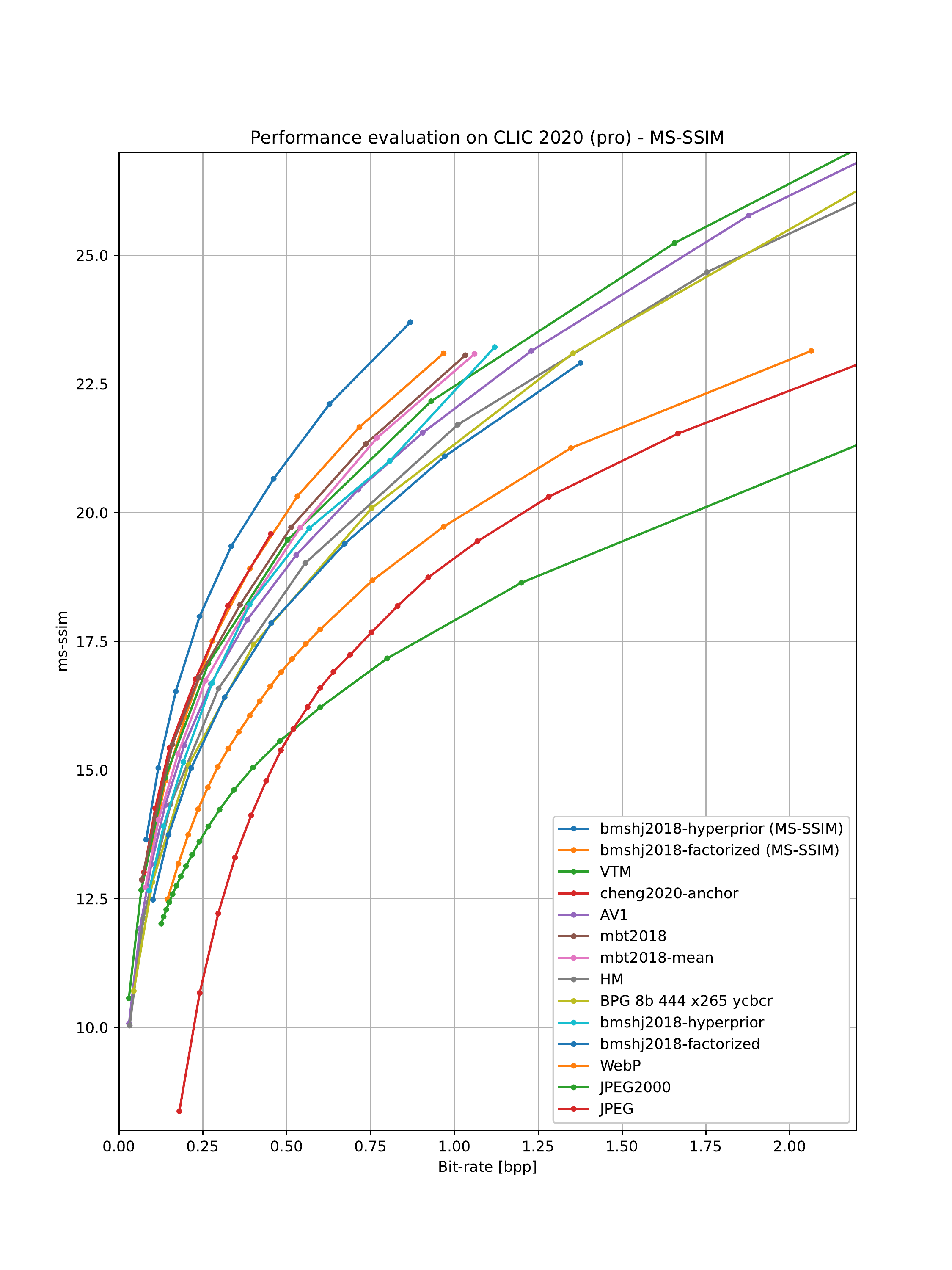}
  \captionof{figure}{\label{fig:results_clic_pro_ms_ssim}%
	  Rate-distortion curves for MS-SSIM measured on the CLIC PRO
  dataset\cite{CLIC}.}
\end{figure}

\newpage
\section{Example Images}
\subsection{Kodak 15}

\begin{figure}[htb]
	\subfloat[Original]
	{\includegraphics[width=0.50\textwidth]{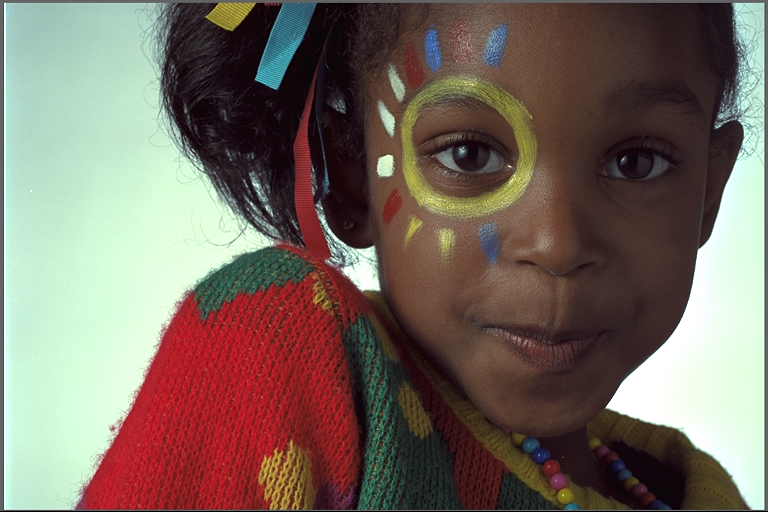}}
	\qquad%
	\subfloat[VVC (0.1156 bpp)]
	{\includegraphics[width=0.50\textwidth]{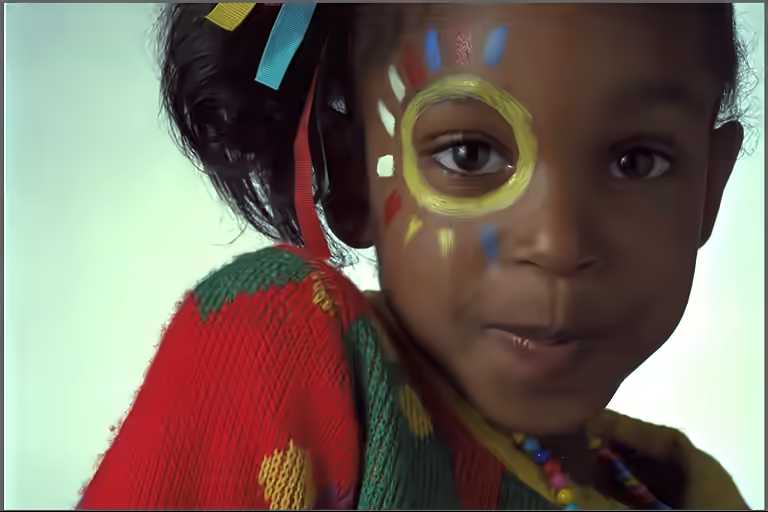}}\\
	\subfloat[cheng2020 anchor (0.1066 bpp)]
	{\includegraphics[width=0.50\textwidth]{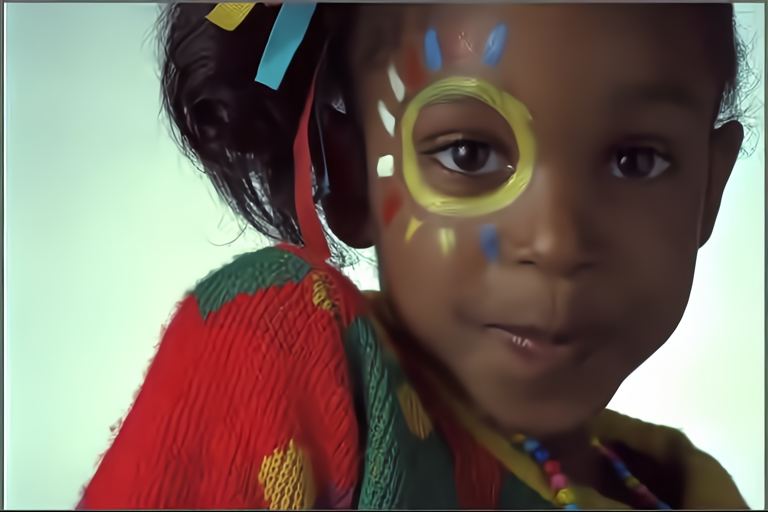}}
	\qquad%
	\subfloat[mbt2018 (0.1068 bpp)]
	{\includegraphics[width=0.50\textwidth]{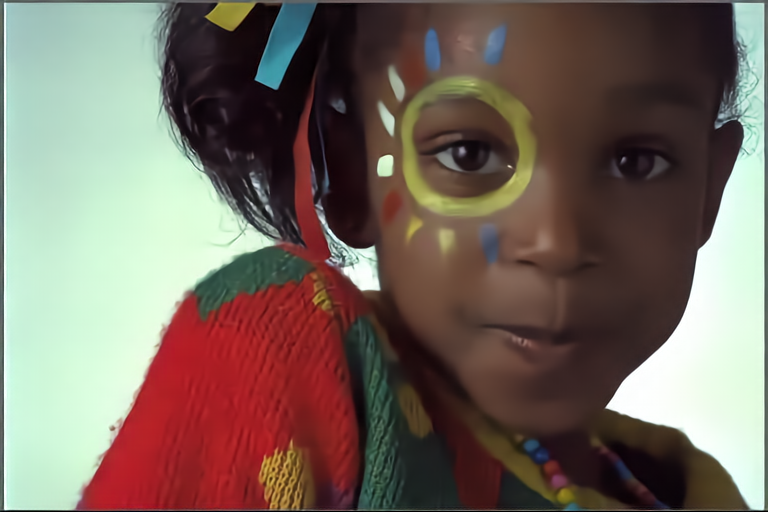}}\\
	\subfloat[BPG (0.1195 bpp)]
	{\includegraphics[width=0.50\textwidth]{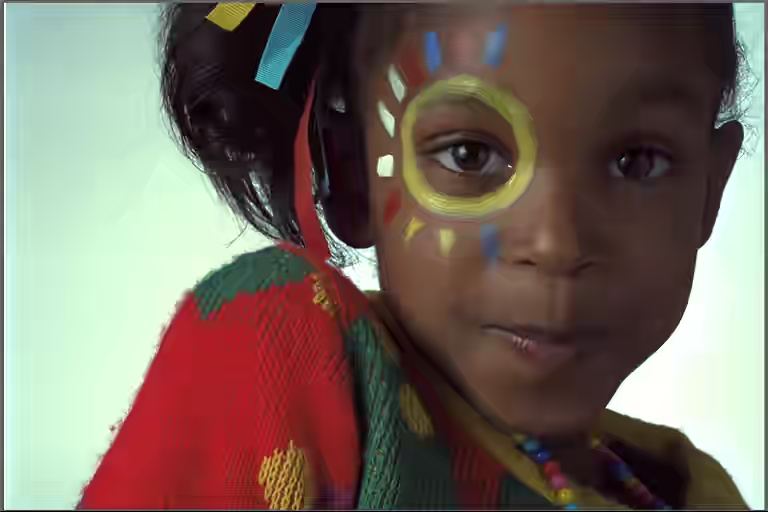}}
	\qquad%
	\subfloat[bmshj2018-hyperprior (0.1391 bpp)]
	{\includegraphics[width=0.50\textwidth]{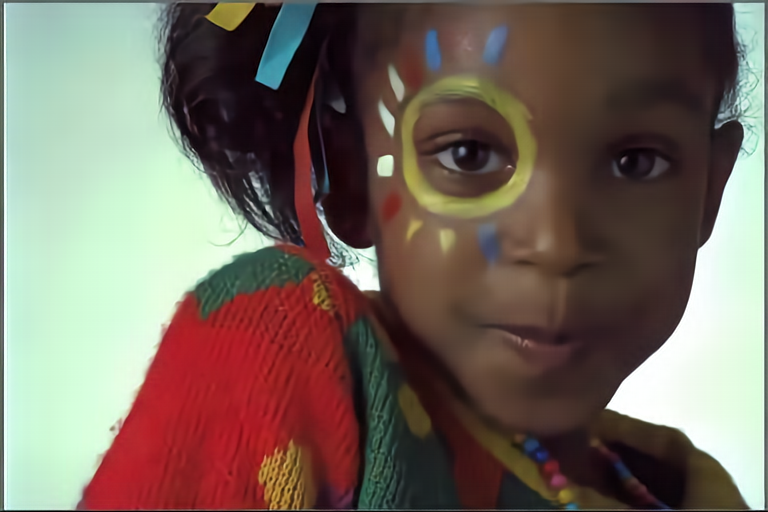}}\\
	\caption{Visual comparison at similar bit-rates for the Kodak 15 image
	(models trained with the mean square error loss).}
\end{figure}

\newpage
\subsection{Kodak 20}

\begin{figure}[htb]
	\subfloat[Original]
	{\includegraphics[width=0.50\textwidth]{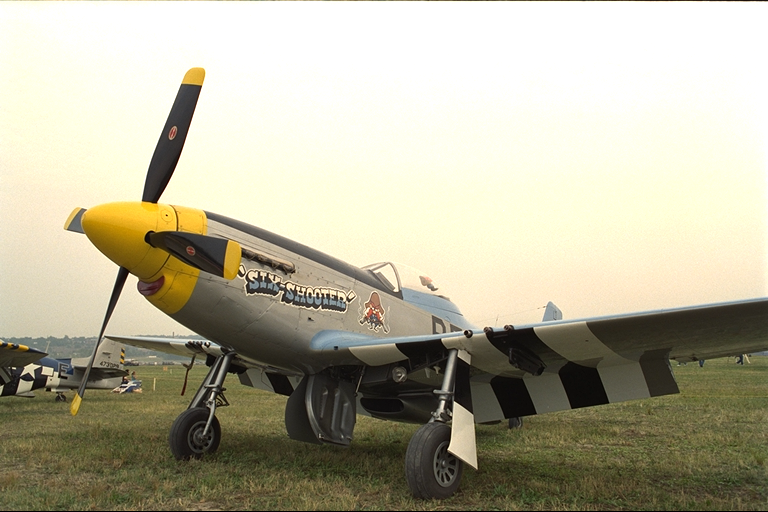}}
	\qquad%
	\subfloat[VVC (0.1099 bpp)]
	{\includegraphics[width=0.50\textwidth]{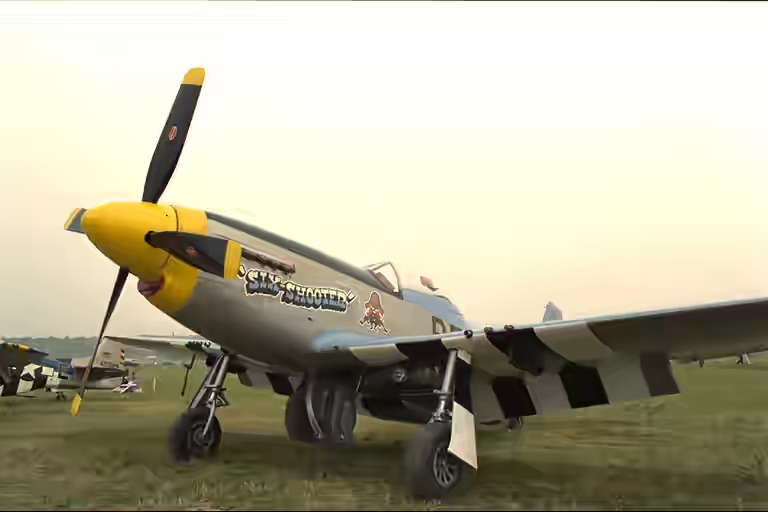}}\\
	\subfloat[cheng2020 anchor (0.1026 bpp)]
	{\includegraphics[width=0.50\textwidth]{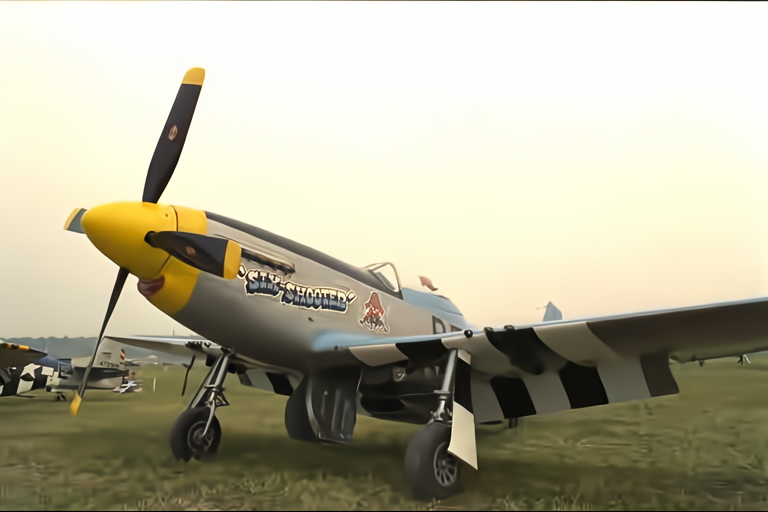}}
	\qquad%
	\subfloat[mbt2018 (0.1018 bpp)]
	{\includegraphics[width=0.50\textwidth]{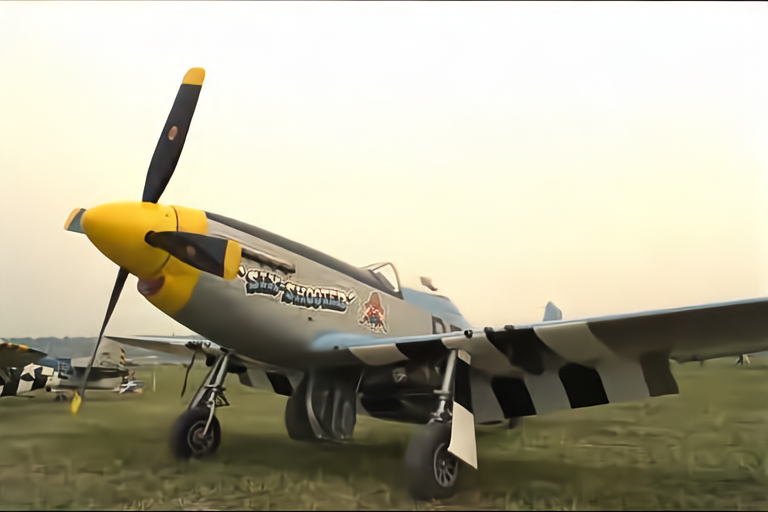}}\\
	\subfloat[BPG (0.1154 bpp)]
	{\includegraphics[width=0.50\textwidth]{assets/imgs/kodim20_bpg_q40.png}}
	\qquad%
	\subfloat[bmshj2018-hyperprior (0.1347 bpp)]
	{\includegraphics[width=0.50\textwidth]{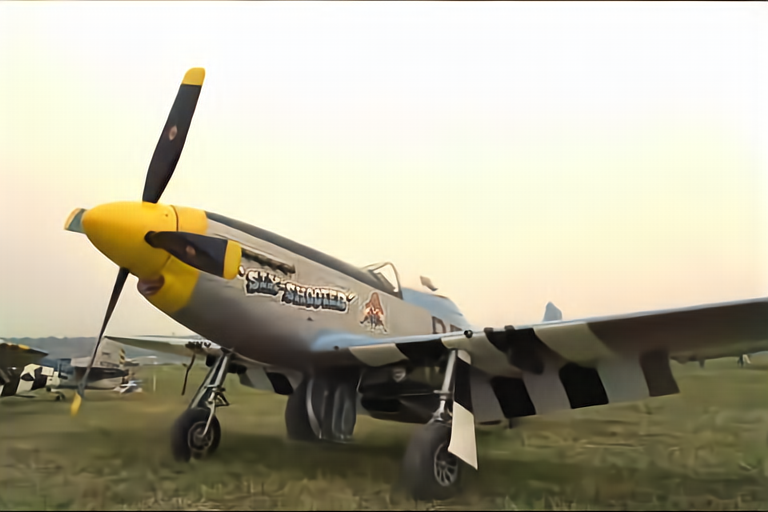}}\\
	\caption{Visual comparison at similar bit-rates for the Kodak 20 image
	(models trained with the mean square error loss).}
\end{figure}

\newpage
\subsection{Saint Malo}

\begin{figure}[htb]
	\subfloat[Original]
	{\includegraphics[width=0.50\textwidth]{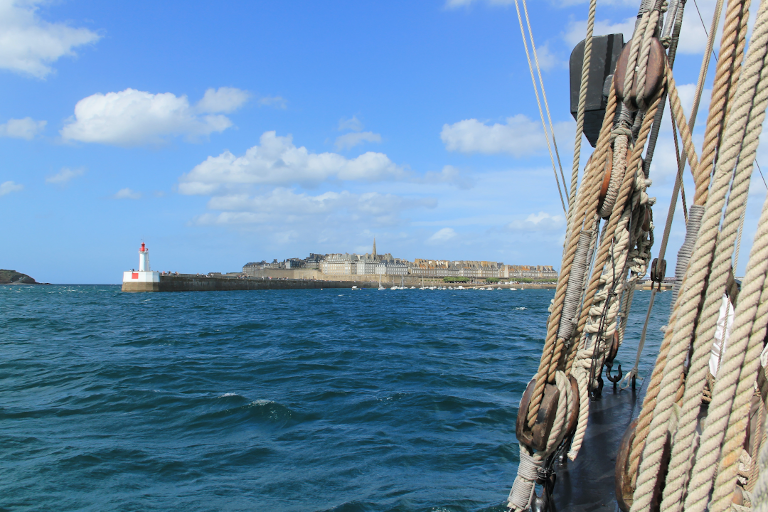}}
	\qquad%
	\subfloat[VVC (0.1763 bpp)]
	{\includegraphics[width=0.50\textwidth]{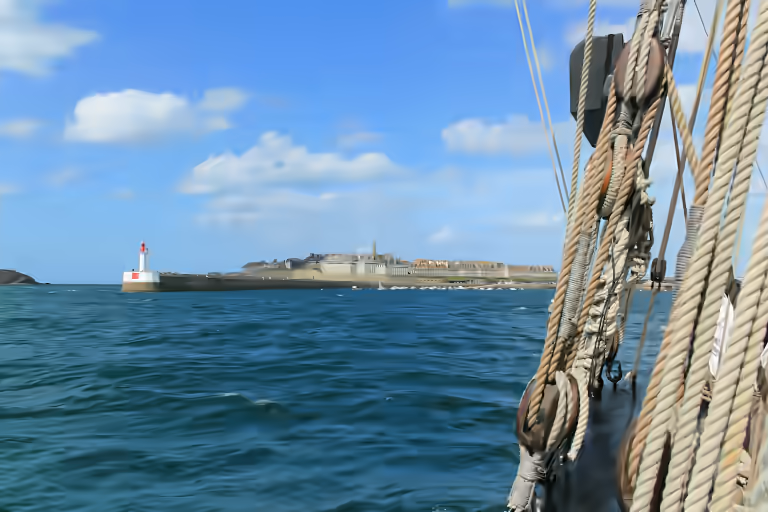}}\\
	\subfloat[cheng2020 anchor (0.2014 bpp)]
	{\includegraphics[width=0.50\textwidth]{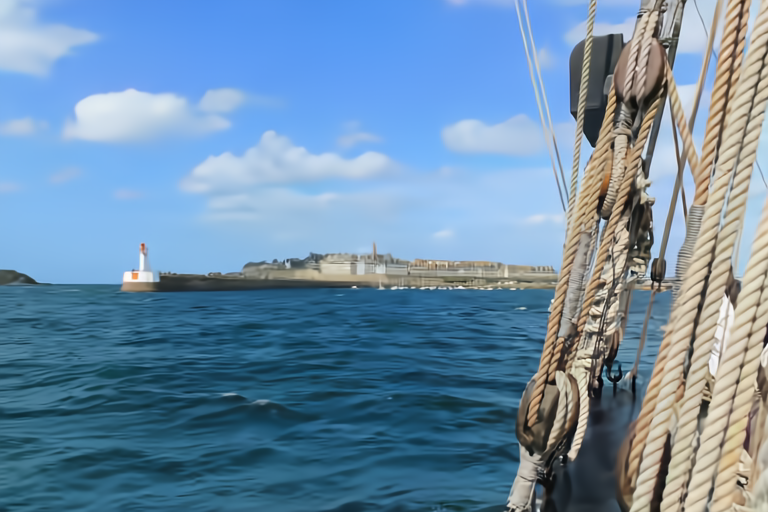}}
	\qquad%
	\subfloat[mbt2018 (0.1976 bpp)]
	{\includegraphics[width=0.50\textwidth]{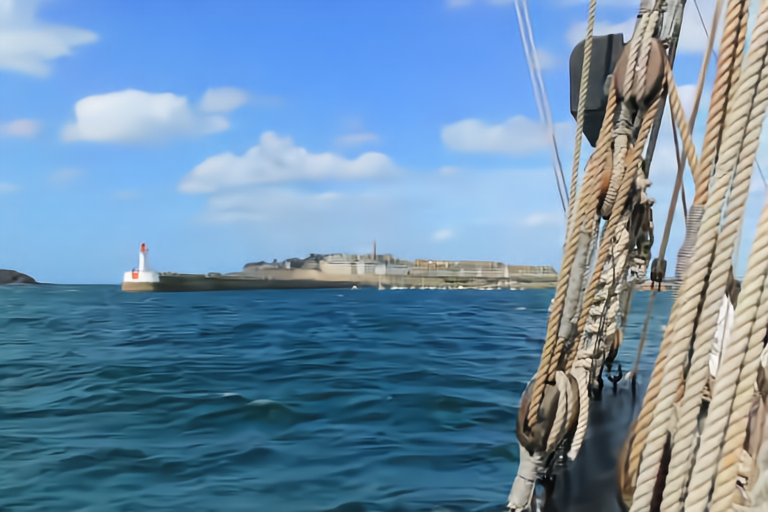}}\\
	\subfloat[BPG (0.1926 bpp)]
	{\includegraphics[width=0.50\textwidth]{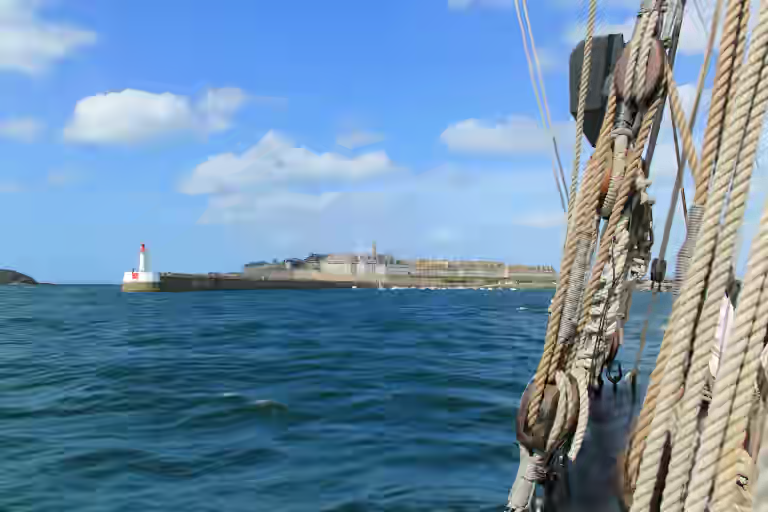}}
	\qquad%
	\subfloat[bmshj2018-hyperprior (0.2260 bpp)]
	{\includegraphics[width=0.50\textwidth]{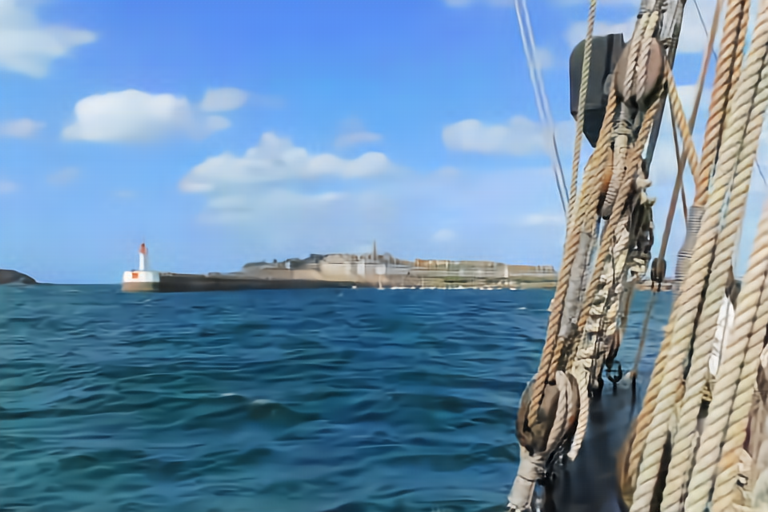}}\\
	\caption{Visual comparison at similar bit-rates for the \textit{Saint Malo}
	image (models trained with the mean square error loss).}
\end{figure}

\section{Code examples}\label{sec:code-examples}

To demonstrate the simplicity of the approach, we include a sample code to build
a simple auto encoder network in Listing~\ref{listing:code-autoencoder}.

The code runs with Python 3.6+, PyTorch 1.5+, Torchvision 0.5 and CompressAI
1.0+. This example model is similar to the fully factorized model presented
in~\cite{balle_end--end_2017}, which is a fully convolutional network with an
entropy bottleneck~\cite{balle_end--end_2017}, and can be replicated in a
few dozens line of codes.

Note that this does not include the code to actually train the network. We
provide an example training code in the examples folder on the CompressAI GitHub
repository\footnote{\url{https://github.com/InterDigitalInc/CompressAI/blob/master/examples/train.py}}.
The full training code is around 300 lines of code, with additional features
such as logging and models check-pointing.

\begin{listing}[ht]
\begin{minted}{python}
import torch.nn as nn

from compressai.entropy_models import EntropyBottleneck
from compressai.layers import GDN

class Network(nn.Module):
    def __init__(self, N=128):
        super().__init__()
        self.encode = nn.Sequential(
            nn.Conv2d(3, N, stride=2, kernel_size=5, padding=2),
            GDN(N)
            nn.Conv2d(N, N, stride=2, kernel_size=5, padding=2),
            GDN(N)
            nn.Conv2d(N, N, stride=2, kernel_size=5, padding=2),
        )

        self.decode = nn.Sequential(
            nn.ConvTranspose2d(N, N, kernel_size=5, padding=2, output_padding=1, stride=2)
            GDN(N, inverse=True),
            nn.ConvTranspose2d(N, N, kernel_size=5, padding=2, output_padding=1, stride=2)
            GDN(N, inverse=True),
            nn.ConvTranspose2d(N, 3, kernel_size=5, padding=2, output_padding=1, stride=2)
        )

   def forward(self, x):
       y = self.encode(x)
       y_hat, y_likelihoods = self.entropy_bottleneck(y)
       x_hat = self.decode(y_hat)
       return x_hat, y_likelihoods
\end{minted}
\caption{\label{listing:code-autoencoder}%
	Example compression network based on the model introduced
	in~\cite{balle_end--end_2017}. More examples are provided in the CompressAI
	Github repository and the online documentation. The code re-implementing the
	state-of-the-art models included in CompressAI is also fully available on
	our repository.}
\end{listing}